\documentclass[review]{elsarticle}

\usepackage{lineno,hyperref}
\modulolinenumbers[5]

\journal{Journal of \LaTeX\ Templates}

\begin{document}

\begin{frontmatter}

\title{Multi-Modality Distillation via Learning the teacher's modality-level Gram Matrix}

\author{Peng Liu}
\address{Yunnan University, Kunming 650500,liupeng0606@gmail.com}

\begin{abstract}
In  the  context  of  multi-modality  knowledge  distillation  research,  the  existing  methods  was  also  mainly  focus  on  the  problem  of  only learning teacher’s final output. Thus, there are still deep differences between the teacher network  and  the  student  network.
It is necessary to force the student network to learn the modality relationship information of the teacher network. 
To effectively exploit transfering knowledge from teachers to students,  a novel modality relation  distillation  paradigm  by  modeling  the  relationship  information  among different modality are adopted,that is learning the teacher’s modality-level Gram Matrix.
\end{abstract}


\end{frontmatter}

\section{Introduction}
With the acquisition of large-scale data and the exploration of deep learning network structure, neural models in recent years have been successful in computer vision and natural language processing field including extremely complex problem statements. For example, UNITER \cite{chen2020uniter}, and BERT \cite{devlin2018bert}. Despite the excellent performance of these networks, there are still problems for application in industry. These networks are huge in size, with millions (and billions) of parameters, most of these existing networks are requires expensive memory and time-consuming computation when inferring, and thus cannot be deployed on edge devices. Although some efficient algorithms have been proposed to solve these problems, the complexity of neural network models is still dramatically increasing, especially in depth. Therefore, research on model compression and inference acceleration for current complex deep neural networks is of great significance. Intuitively, the theoretical search space for more complex models is larger than for a smaller network. The convergence space of the larger network should therefore overlap with the solution space of the smaller network if the same (or even similar) convergence can be achieved with a smaller network. However, the smaller network usually cannot converge by itself. Smaller networks often have differences in convergence from larger networks. In contrast, if the smaller network is guided to replicate the behavior of the larger network, then the smaller network convergence space is likely to overlap that of the larger network. On the other hand, Previous studies \cite{breiman1996born} 
\cite{ba2013deep}
\cite{huang2017like}

\cite{hinton2015distilling} have also suggested that deep neural networks often have redundancy, so it is feasible to compress the model.
In recent years, there have been several proposed methods for compressing neural networks. These methods are generally categorized into three major categories: Quantization, Weights Pruning, and Knowledge Distillation (KD). In these, KD has received much success for reducing the size of pretrained large models. Knowledge distillation was first proposed by Bucilu et al. in 2006. In these study, they tried to transfer the output of a large network to a shallow network. Later, hinton redefines knowledge distillation, which refers to the idea by teaching a smaller network, step by step, exactly imitating a bigger already trained network output. Knowledge Distillation assumes that the knowledge learned by the teacher is a mapping from input to output, during the training process, the output of the last layer of the teacher is passed to the students as the goal.

Knowledge distillation can transfer the knowledge of one network to another. The two networks can be isomorphic or heterogeneous. The method is to train a teacher network first, and then use the output of the teacher network and the real label of the data to train the student network. Knowledge distillation can be used to transform a network from a large network to a small network, and retain the performance close to that of a large network; The knowledge learned from multiple networks can also be transferred to one network, so that the performance of a single network is close to the result of emsemble.

Despite knowledge distillation has explored in various studies \cite{furlanello2018born} \cite{yim2017gift} 
\cite{zagoruyko2016paying}
\cite{yang2018learning}
\cite{schroff2015facenet}
\cite{kim2018attention}
in recent years, such as improving student models and improving teachers performance via self-distillation, there are still not involved in the intensive study of the multi-modal distillation, which are even rarely involved in prior works. For example, the visual entailment (VE) task, where the premise is defined as an image rather than a natural language sentence relative to traditional textual entailment (TE). It contains text and image information, and the respondent needs to judge the relationship between the text and the image (i.e. Entailment, Neutral and Contradiction). Research involving multi-modality is of great significance, because in the real world, some information is more often presented in a combination of images and text. This requires deep learning models to integrate and understand multi-modal information well. Although the existing knowledge distillation methods can be applied to multi-modal distillation, the student network directly learns the output of the teacher network. However, it ignores the rich relationship information of different modality in the teacher network. For example, the relationship information of teacher’s output when images and text are separately input to teacher's network.

Unfortunately, in the context of multi-modality knowledge distillation research, the existing methods was also mainly focus on the problem of only learning teacher’s final output. xx’s research suggested that only distillating the multi-modal knowledge of the teacher network will have a major disadvantage: for per modal information, there are still deep differences between the teacher network and the student network. Therefore, some knowledge of the teacher network cannot be effectively transferred to the student network only through the method of conventional knowledge distillation. Thus, for the multi-modal teacher model, it is necessary to force the student network to learn each modality and modality relationship information of the teacher network. 
Different from the previous methods, we decided to explore the relationship between different modality. Inspired by this method, we design a novel modality relation-driven  framework for Multi-Modality Distillation. As shown in Figure 1, To effectively exploit transfering knowledge from teachers to students, a novel modality relation Distillation paradigm  by modeling the relationship information among different modality are adopted.

Our main contributions:
We not only explore the final output of multi-modal knowledge distillation as the only distillation goal, but our method can effectively explore transfering information between different modality corresponding to teachers to students, so as to improve the performance of multi-modal distillation.

\section{Related works}
\subsection{Knowledge Distillation}
With the rapid increases in computing power, it is not surprising that various complex deep neural networks with a large number of parameters such as UNITER \cite{chen2020uniter}, and BERT \cite{devlin2018bert} have been increasingly used for computer vision natural and language processing and have achieved great success \cite{romero2014fitnets}
\cite{zagoruyko2016paying}
\cite{bagherinezhad2018label}. However, it can not be efficiently deployed on device with limited computing and storage capability \cite{cao2018fast}\cite{krizhevsky2009learning} \cite{vapnik2015learning}. To address the above issues, research mainly focuses on model compression such as knowledge distillation \cite{sau2016deep}, \cite{polino2018model}, model quantization \cite{polino2018model}, \cite{zhou2018adaptive} \cite{fan2020training} \cite{hansen2014relativistic} and model pruning \cite{liu2018rethinking}, \cite{zhu2017prune}. Among them, the knowledge distillation approach has been widely used due to its advantages, such as low performance sacrifice, easy implementation and hardware-friendly.
The vanilla knowledge distillation involves training a small model (student) to match a large pre-trained model (teacher). In order to transfer the knowledge from the teacher model to the student, a loss function is optimized to match ground-truth labels as well as softened teacher logits. 
As the "temperature" scale function is applied to the softmax, the logit distributions learn by teacher become softer, which can reveal inter-class relationships effectively.
Based on intuition, the key to the success of KD is mainly that more fine-grained supervised information for improving the student model performance are provided across different categories in soft targets rather than discrete labels. Unlike previous interpretations, new concepts propose that soft target regularization functions as smoothing regulation for preventing overconfident predictions by student models.

In recent research \cite{zhu2021student} \cite{wang2021knowledge} \cite{panchapagesan2021efficient} \cite{chen2021simplified}
\cite{shang2021lipschitz}
\cite{liu2021semantics}
\cite{zhao2021novel}, the main goal of knowledge distillation is to transfer the feature information of samples from teachers to students. For example, \cite{sen2009meta} imitated the teacher network by asking students to learn to return Logits before the softmax layer. \cite{phuong2019towards} let students share some lower semantic levels with teachers and train them at the same time, but they also let students learn teachers' Logits knowledge. In order to transmit the middle layer information learned by the teacher network from the sample to the student network, \cite{mirzadeh2020improved} proposed fitnet, which uses the feature mapping and final output of the middle layer of the teacher network to teach the student network. However, these methods have a common feature. They only learn the feature information of a single sample from the teacher network, and rarely test the relationship between sample features. In addition, the characteristics of teacher network middle layer are closely related to the actual situation Network design, so the above methods can not be widely popularized. In addition, most methods directly force students to learn the output of teachers' network, while ignoring the feature space transformation process \cite{huang2021revisiting} \cite{xu2021kdnet}. In order to solve this problem, \cite{yim2017gift} proposed the solution process (FSP), which is designed to let students learn online and teachers learn, rather than the results of the middle layer.

Different from their methods, we do not explore how to transfer the relationship information between samples from teacher network to student network.
Our goal is to explore how to transfer the relationship information of different modality from teacher network to student network.

\begin{figure*}[!t]
\centerline{\includegraphics[width=7in]{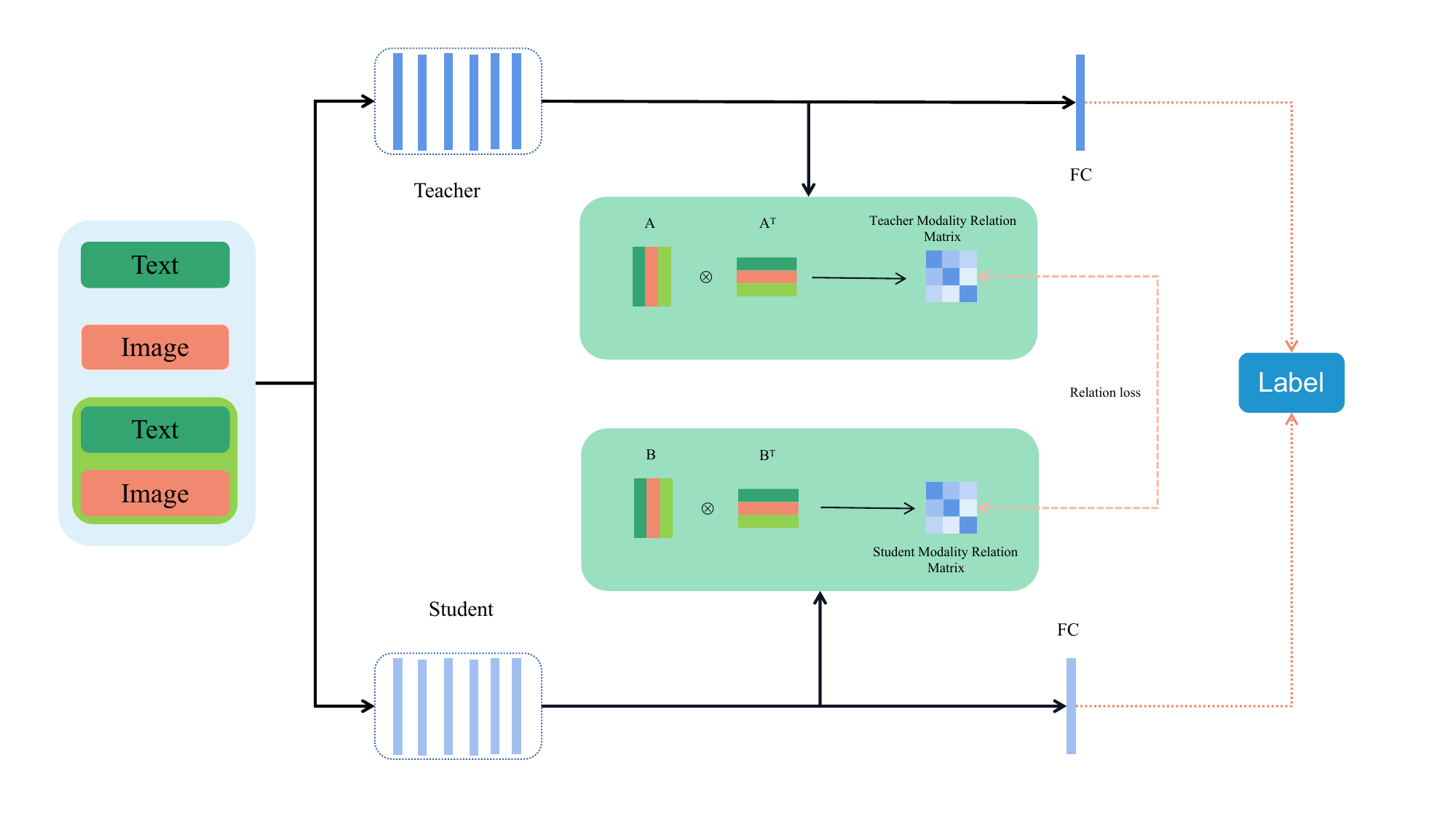}}
\caption{The detail architecture diagram of our method.}
\label{fig1}
\end{figure*}

\section{Our Approach}

In our method, there are two parts of loss, that is, the traditional KD loss and our proposed modality relationship loss between teacher and student. We first introduce the KD loss, and then introduce our proposed loss .

Generally speaking, the traditional KD knowledge distillation \cite{aguilar2020knowledge}
\cite{aguilar2020knowledge}
\cite{xu2020knowledge}
\cite{wang2018kdgan}
\cite{tang2020understanding}
\cite{liu2019knowledge}
\cite{mishra2017apprentice}
can be viewed as minimizing the objective function:

\begin{equation}
\mathcal{L}_{\mathrm{IKD}}=\sum_{x_{i} \in \mathcal{X}} l\left(f_{T}\left(x_{i}\right), f_{S}\left(x_{i}\right)\right)
\end{equation}
where the $l$ is represent the loss function, which is used to penalize the difference between teacher network and student network.

In our paper, we input three modal information to teacher and student network in turn, that is, text information alone, picture information alone, and joint information of picture and text. For the student network output of these three modals, we adopt the standard cross entropy loss between student outputs and ground true label, which can be regarded as a data enhancement strategy, which can improve the performance of the model. This can be defined as follow:

\begin{equation}
\mathcal{L}_{\mathrm{CE}}= \alpha CE\left(f\left(x_{t}\right),y) + \beta CE(f\left(x_{i}\right), y) + \gamma CE(f\left(x_{i+t}\right), y)  \right) 
\end{equation}

where the $x_{i}$ represent the image modality, the $x_{t}$ represent the text modality, the $x_{t+i}$ represent the text and image modality.

Inspired by recent research, we can learn extra semantic information
for an entity,  Based on this ideas, we propose to model such modality relation to transfer knowledge from teacher to student.
Thus, our method aims at transferring the relationship knowledge of different modality  using mutual modality relations in the teacher’s output.

We model the modality relationship in a single sample
with a modality-level Gram Matrix [42]. Given an input sample
that can be divided into three modality (image, text, text and image), 
We denote the output results when the network inputs these three modal information as $A$, thus, for a single sample, the modality relationship $G$ can be defined as follow:
\begin{equation}
G = A \cdot A^{T}
\end{equation}

Our goal is to transfer teacher's modality relationship to student, which can be defined as follow:

\begin{equation}
\mathcal{L}_{mr}= MSE\left( A_{t}, A_{s} \right)
\end{equation}

where the MSE represent the loss function of mean square error, the  $A_{t}$ and $A_{s}$ represent the information of teacher and student modality relationship.

\section{Experiment}
We evaluated our proposed multimodal distillation method, in this section, we will describe the experiment in detail.
\section{Datasets}
To demonstrate the effectiveness of our approach, we pick up three multimodal
datasets, including Hateful-Memes, SNLI-VE, and NLVR.
The Hateful-Memes dataset consists of 10K multimodal memes. The task is a binary classification problem, which is to detect hate speech in multimodal memes. We use Accuracy (ACC) as evaluation metrics for hateful memes.
The goal of Visual Entailment is to predict
whether a given image semantically entails an input sentence. Classification accuracy over three classes ("Entailment", "Neutral" and "Contradiction") is used to measure model performance. We use accuracy as an evaluation metric following.
NLVR contains 92,244 pairs of human-written English sentences grounded in synthetic images. Because the images are synthetically generated, this dataset can be used for semantic parsing.

\section{Implementation details}
For the teacher model, we use a 12 layer pre-trained uniter network, and for the student model, we use a 2-layer of uniter network to implementation student network. In our task, we only consider the relationship between image and text. Conventional KD was used as the basic distillation method in this paper. In addition, we include several distillation method baselines including conventional KD. Other distillation methods are also applicable to our method and we will discuss the results in our experiments using other KD methods. For analysis, we used UNITER, pre-trained multimodal models such as teacher model and 2 layer UNITER pre-trained multimodal models as a student model. UNITER consists of 12 layers with a hidden size of 768. 
Student model consists of 2 layers with a hidden size of 768. We used the regional features in the images as a kind of fine tuning for both the teacher and the student on each dataset for the student. Validation sets used to train weight learners to use data sets as metadata data. We find the optimal hyperparameter on the validation set.
\textbf{\begin{figure*}[]
\centerline{\includegraphics[width=5in]{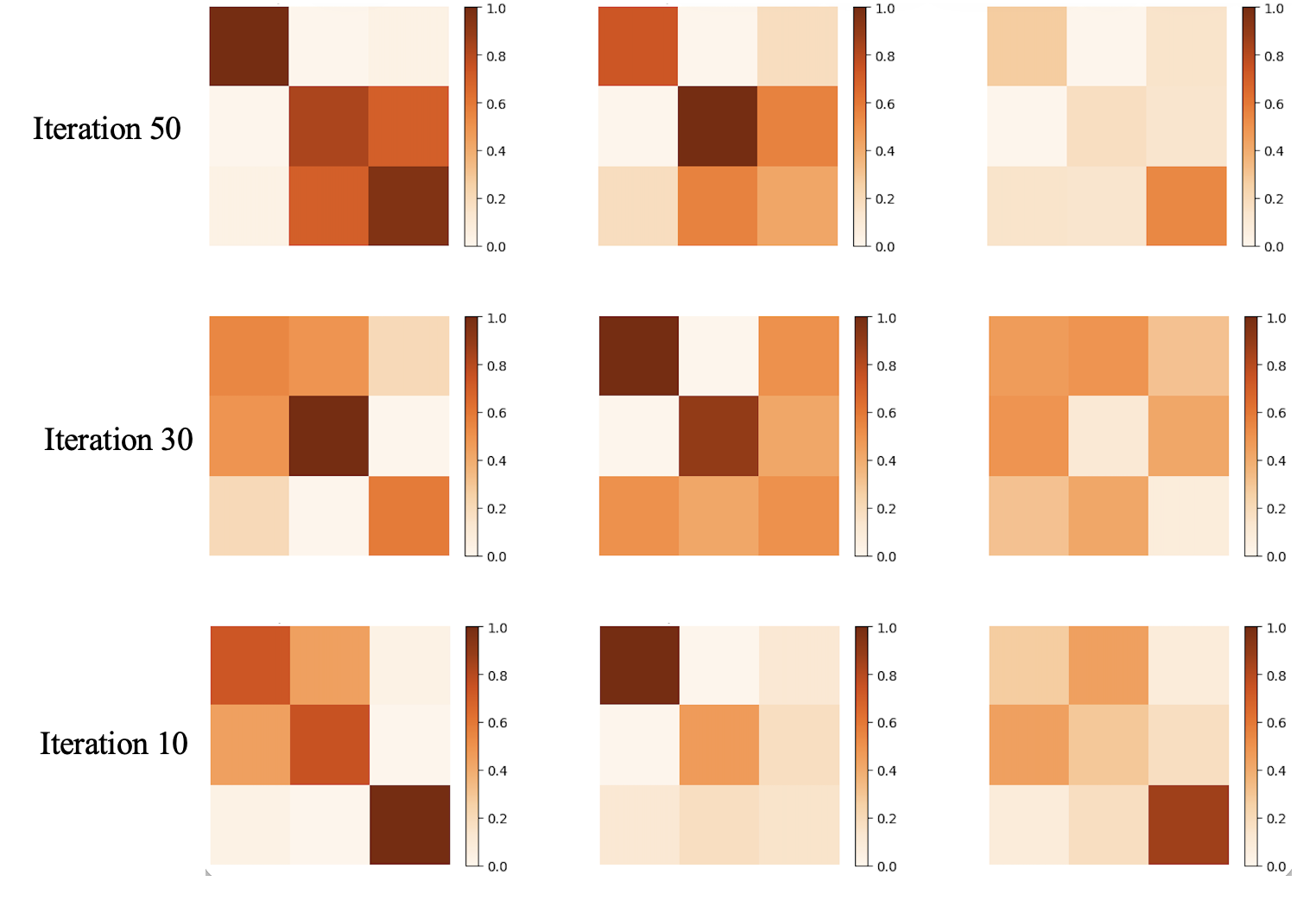}}
\caption{The modality relationship result of our method.}
\label{fig1}
\end{figure*}}

\section{Experiment result}

\subsection{Compare our method with others}
Our method in detail as shown in table 1 result, you can see our way to a baseline method on the basis of the relative to a promotion, it may be that the way for us to learn from teachers in the network to more information for classification, the way we design, you can learn from teachers in the network to a single sample of the interaction between the different modal information, This information is effectively transferred from the teacher network to the student network. Our approach has better performance than just using the last layer of output information.

\begin{table}[]
\setlength{\tabcolsep}{3mm}
\renewcommand\arraystretch{1.5}
\begin{tabular}{ccccccccc}
\hline
                          & \multicolumn{2}{c}{VE} &  & \multicolumn{2}{c}{NLVR} &  & \multicolumn{2}{c}{HM} \\ \cline{2-9} 
                          & test       & val       &  & test        & val        &  & test       & val       \\ \hline
\multicolumn{1}{c|}{KD}   & 71.22      & 71.43     &  & 73.62       & 73.45      &  & 68.22      & 67.89     \\ \hline
\multicolumn{1}{c|}{Ours} & 72.45      & 72.66     &  & 75.33       & 75.06      &  & 69.54      & 69.85     \\ \hline
\end{tabular}
\caption{the result of comparing our method with others}
\end{table}

\subsection{Learning relationship matrix from teacher}
In order to better understand the learning behavior of our paradigm in network training, we visualized the sample relationship matrix $G$ after normalization, as shown in Fig 2, from the development of student model and teacher model in different training times. To clearly show the alignment of the two matrices, we also compute their absolute distance matrices, as shown in the red column to the right. As can be seen from Fig.2, at the beginning of the network training, the internal connection structure of different samples was not well presented (note that small batches were classified according to the basis truth label), and the calculated relationship matrix was greatly different due to input disturbance. As the training progressed, the model gradually generates meaningful relationship matrix, and the matrix of the student network and the matrix of the teacher network are more and more similar. At the same time, with the convergence of the model, the absolute difference between the two becomes smaller and smaller, indicating that the student model has gradually learned the modal information.

\section{Conclusion}
In this paper, we propose a novel method for  multi-modality  knowledge  distillation,  while the  existing methods was also mainly focus on the problem of only learning teacher’s final output.  Thus, there are still deep differences between the teacher network and the student network.  It is necessary to force the student network to learn the modality  relationship  information  of  the  teacher  network.   To  effectively  exploit transfering knowledge from teachers to students, a novel modality relation Distillation paradigm by modeling the relationship information among different modality are adopted,that is learning the teacher’s modality-level Gram Matrix.

\bibliography{paper}

\end{document}